# Bridging the Knowledge-Action Gap by Evaluating LLMs in Dynamic Dental Clinical Scenarios

Hongyang Ma[1, 2#], Tiantian Gu[3#], Huaiyuan Sun[3#], Huilin Zhu[2, 4#], Yongxin Wang[3#], Jie Li[1], Wubin Sun[3], Zeliang Lian[3], Yinghong Zhou[4], Yi Gao[5*], Shirui Wang[3*], Zhihui Tang[2*]

[1]The Second Affiliated Hospital of Harbin Medical University, Heilongjiang Province, China
[2]Peking University School and Hospital of Stomatology, Beijing, China
[3]Medlinker Intelligent and Digital Technology Co., Ltd, Beijing, China
[4]School of Dentistry, Centre for Orofacial Regeneration, Rehabilitation and Reconstruction (COR3), The University of Queensland, Brisbane, Australia
[5]Department of Stomatology, Beijing Xicheng District Health Care Hospital for Mothers and Children, Beijing, China

*Corresponding authors:

Yi Gao (13439023411@126.com),

Shirui Wang (wsr@medlinker.com),

Zhihui Tang (zhihui_tang@126.com),

#These authors contributed equally to this work

## Abstract

The transition of Large Language Models (LLMs) from passive knowledge retrievers to autonomous clinical agents demands a shift in evaluation—from static accuracy to dynamic behavioral reliability. To explore this boundary in dentistry, a domain where high-quality AI advice uniquely empowers patient-participatory decision-making, we present the Standardized Clinical Management & Performance Evaluation (SCMPE) benchmark, which comprehensively assesses performance from knowledge-oriented evaluations (static objective tasks) to workflow-based simulations (multi-turn simulated patient interactions). Our analysis reveals that while models demonstrate high proficiency in static objective tasks, their performance precipitates in dynamic clinical dialogues, identifying that the primary bottleneck lies not in knowledge retention, but in the critical challenges of active information gathering and dynamic state tracking. Mapping "Guideline Adherence" versus "Decision Quality" reveals a prevalent "High Efficacy, Low Safety" risk in general models. Furthermore, we quantify the impact of Retrieval-Augmented Generation (RAG). While RAG mitigates hallucinations in static tasks, its efficacy in dynamic workflows is limited and heterogeneous, sometimes causing degradation. This underscores that external knowledge alone cannot bridge the reasoning gap without domain-adaptive pre-training. This study empirically charts the capability boundaries of dental LLMs, providing a roadmap for bridging the gap between standardized knowledge and safe, autonomous clinical practice.

# Introduction

In recent years, the trajectory of Large Language Models (LLMs) in healthcare has rapidly shifted from passive information retrieval to the development of autonomous clinical agents capable of managing critical stages, including history taking, diagnostic suggestions, treatment planning, and follow-up management. Notably, the emergence of specialized multimodal models, such as DentVLM, has demonstrated the potential to elevate junior practitioners' diagnostic performance to expert levels and significantly reduce diagnostic time by integrating visual and linguistic information[1]. Concurrently, comprehensive evaluations indicate that LLMs perform strongly in clinical note generation and patient communication[2]. These capabilities align naturally with dentistry, a field characterized by its predominantly outpatient nature, highly standardized workflows, and frequent interdisciplinary collaboration, making it an ideal testbed for agentic deployment[3]. Beyond operational efficiency, the deployment of AI agents in this field holds profound significance for patients: it enables them to obtain high-quality diagnostic and treatment recommendations directly via online platforms. This accessibility breaks down information barriers, empowering patients to truly participate in clinical decision-making alongside professionals.

To rigorously assess the foundational competencies required for such agents, the academic community has established a series of benchmarks evolving from static knowledge to dynamic reasoning. Foundational datasets like MedQA (USMLE-based) and MedMCQA have served as the gold standard for evaluating the "knowledge brain" of potential agents through large-scale multiple-choice questions[4, 5]. Moving beyond rote memorization, benchmarks such as PubMedQA have been introduced to evaluate models' abilities to reason over biomedical literature and evidence, a crucial skill for evidence-based practice[6]. Furthermore, leading research initiatives, such as the MultiMedQA suite used to evaluate Med-PaLM, have aggregated these datasets to set new standards for "expert-level" performance in general medicine[7].

However, a critical "Knowledge-Action Gap" remains unaddressed in current evaluation paradigms. Existing systems predominantly focus on general Q&A or single-point tasks, failing to capture the performance degradation that occurs when models transition from selecting static options to managing dynamic patient interactions[8]. Consequently, two critical questions regarding the feasibility of dental agents remain unresolved: First, does the LLM possess a robust "knowledge brain" that is factually accurate? Second, can this knowledge be safely translated into "clinical execution" within a multi-turn workflow?

Regarding the first question—knowledge retention versus application—literature reveals a paradox of "high exam scores versus limited practical capability." Systematic reviews indicate that high-performance models like GPT-4 have achieved accuracy rates in dental licensing examinations that surpass the average human candidate[9, 10]. Furthermore, in the domain of educational content production, board-style questions generated by LLMs have achieved difficulty and discrimination indices comparable to those created by human experts[11]. Yet, these "high scores" do not directly translate to reliable agentic competence. Studies highlight that once tasks involve open-ended clinical inquiries, significant disparities emerge, often accompanied by "confident but wrong" hallucinations[12-14]. This necessitates a re-evaluation of the "knowledge brain" not just by score, but by its stability across guidelines and languages.

Regarding the second question—the balance between safety and efficacy in execution—current evaluations reveal the limitations of pursuing "accuracy" alone. In clinical practice, safety is embodied in identifying contraindications

(e.g., antibiotic prophylaxis for infective endocarditis), while efficacy focuses on regimen optimization. The CRAFT-MD framework[15], found that even diagnostically accurate models exhibit critical limitations in "conversational reasoning," often failing to ask follow-up questions to elicit key information. Such passivity poses unacceptable risks in real-world flows, where performance is often unstable and dependent on prompt engineering[16]. Furthermore, while specific models demonstrate strong reasoning in complex implantology cases, the risk of misleading advice remains high without a mechanism to define safe operating boundaries[17]. This suggests that current assessments have not yet sufficiently decomposed and weighted "safety-critical points" against "efficacy optimization points."

To bridge this gap, this study presents the Standardized Clinical Management & Performance Evaluation (SCMPE) benchmark. Addressing the linguistic bias inherent in mainstream benchmarks[9, 18-21], we construct a bilingual (Chinese-English) evaluation protocol. Drawing on the interactive concepts of CRAFT-MD and the hallucination-focus of Med-HALT, we seek not only to quantify the "Knowledge-Action Gap" by contrasting objective exam performance with multi-turn dialogue scores, but also to map the decision boundaries of LLMs using fine-grained safety and effectiveness metrics, and to validate the impact of Retrieval-Augmented Generation (RAG) in optimizing these boundaries. This approach intends to provide empirical evidence with greater external validity, facilitating the transition of LLMs from "exam high-achievers" to "reliable clinical copilots".

## Methods

### Data Acquisition and Construction of the Bilingual Repository

To empirically chart the capability boundaries of LLMs in dentistry, we first engineered a comprehensive bilingual repository designed to stress-test clinical-agent readiness from knowledge-oriented evaluations to workflow-based simulations, under a unified clinical-alignment rubric. This repository was curated from three distinct sources to ensure coverage from theoretical foundations to real-world complexities. For the assessment of static memory and reasoning, we aggregated a vast collection of Bilingual MCQs from the MedMCQA[5] dental subset and the Chinese National Dental Licensing Examination database. Following a rigorous filtration process to remove non-dental entries and verify explanation logic, we retained 2,192 English and 2,066 Chinese items. To evaluate evidence-based retrieval capabilities, we compiled 60 authoritative clinical practice guidelines, comprising 19 Chinese and 41 English documents. For workflow-based simulation aimed at approximating real-world deployment, we collected 118 Real-world Case Reports (68 Chinese, 50 English) covering diverse dental sub-specialties. Notably, to minimize data contamination, the collected clinical guidelines and real-world case reports were not sourced from public general datasets or common internet medical corpora. This reduces the likelihood of their presence in the LLMs' pre-training data, ensuring the evaluation targets generalization capability rather than rote memorization. These raw data sources (eTable 1) formed the foundational substrate for our subsequent structured annotation.

### Establishment of Clinical Alignment Criteria via Delphi Consensus

To transform raw medical data into a rigorous evaluation standard, we defined the operational boundaries and safety guardrails required for an autonomous agent through a three-round Delphi consensus process. The objective was to establish a "Clinical Alignment" framework that quantifies the consequence of agentic error. The initial protocol proposed 39 distinct performance indicators. In each round, an expert panel provided retention judgments, quantitative criticality weights, and qualitative refinement proposals. We synthesized feedback between rounds, integrating statistical stability indicators to iteratively refine the agent's compliance requirements. The process converged after three rounds, crystallizing into 37 validated alignment metrics. Based on the consensus median scores, we instantiated a Risk Stratification Framework that maps each metric to a 1–5 severity scale. This scale delineates the impact of error, ranging from Weight 1 (Context-dependent/limited risk) to Weight 5 (Life-threatening or catastrophic failure), thereby providing a weighted ground truth for the subsequent scoring of agent behaviors (detailed in eTable 2). High-risk and emergency topics achieved the highest scores with stable consensus: "Emergency management of airway obstruction," "Contraindications for extraction," "Bone volume assessment for implants," "Decision for tooth preservation vs extraction," "Radiographic assessment for impacted tooth extraction," and "Indications assessment for implant surgery" fell within [4.5,5] (score 5). Items such as "Drug–drug interaction alerts," "Early signs of LAST," and "Informed consent essentials" fell within [4,4.5) (score 4). Some effectiveness or foundational diagnostic items (e.g., "Guideline-based differential diagnosis," "Accuracy of caries grading," "Systemic–oral interaction risk assessment") had medians in [0,3) or [3,3.5), scoring 1–2. Most items achieved IQR≤1 by rounds two to three, with the consensus median as the final representative weight, indicating good convergence and clear prioritization.

## Structured Annotation and Test Set Generation

Leveraging the 37 validated alignment metrics, we employed a "Human-in-the-loop" knowledge engineering workflow to convert the raw repository into a structured evaluation suite. For knowledge-oriented evaluations (Guideline-based Open QA), the collected guidelines underwent semantic chunking and were mapped to the consensus checkpoints. We utilized Qwen (qwen3-max-preview) to draft initial QA pairs per chunk, which then underwent rigorous clinician review to ensure the reference answers aligned strictly with the 37 consensus metrics. This yielded 205 complex queries covering 443 checkpoint instances and 1,225 rubric points. For workflow-based simulations (simulated deployment based on real-world cases), the real-world case reports were destructured into clinical facts (case vignette) and re-mapped to the consensus metrics to create high-fidelity simulation scenarios. All mappings were clinician-verified, resulting in a dataset of 498 checkpoint instances and 1,318 rubric points. This rigorous annotation process ensured that every evaluation point in the test set was directly traceable to the expert-defined safety and effectiveness standards (eTable 3).

## Experimental Setup: Multi-turn Dialogue Simulation

To assess the feasibility of autonomous deployment, we configured an experimental environment that mirrors the dynamic nature of clinical practice. We deployed Qwen as a Simulated Patient (SP) LLM to engage the target LLM in multi-turn dialogues. The SP LLM ingested the structured case vignette and operated under strict prompt constraints to prevent role-breaking or information leakage. The target LLM was tasked with executing a complete clinical workflow—from history taking to iterative planning. The target LLM was instructed to autonomously terminate the dialogue only when it had gathered sufficient information and was 100% confident in the diagnosis (see Prompts in Supplementary Note 1), subsequently outputting a structured management plan. This mechanism forces the model to judge information saturation, thereby directly evaluating its active history-taking and information-gathering capabilities.

Furthermore, to deconstruct the capability gap between "inquiry" and "reasoning," we designed a control experiment. In this setting, the full Case Vignette held by the Simulated Patient (SP) was directly fed into the target LLM, allowing it to bypass the inquiry phase and generate a management plan immediately. By comparing scores between the "Multi-turn Inquiry Mode" and the "Direct Input Mode," we can isolate performance deficits caused by information gaps due to poor inquiry strategies from those stemming from intrinsic flaws in clinical reasoning logic.

**Unified Scoring Mechanism: Safety Guardrails versus Clinical Effectiveness**

We implemented a unified rubric-based scoring framework tailored to the dual nature of our evaluation metrics. For Safety Checkpoints, we adopted a "zero-tolerance" principle to evaluate the agent's adherence to critical guardrails. This was operationalized via a binary constraint mechanism where a score of 1 is awarded only if pass criteria are met and no fail criteria are triggered; otherwise, the score is 0. For Effectiveness Checkpoints, we utilized a dynamic scoring scheme to measure the quality of clinical execution. Each checkpoint contained multiple rule-based criteria weighted from -10 to 10. These scores were aggregated and normalized to a range of [0, 1] at the checkpoint level. While the weighted normalized total score reflects the model's macro-logical capability in handling clinical tasks, we emphasize the independence of Safety Guardrails. A safety score of 0 serves as a "clinical veto," indicating that regardless of how high the efficacy score is, the model poses an unacceptable risk in that scenario. This dual-scoring approach allows for a granular analysis of the trade-off between "Guideline Adherence" (Safety) and "Clinical Decision Quality" (Effectiveness).

**Native-Language Evaluation and Regional Alignment**

Distinct from standard multilingual benchmarks that rely on parallel translation, our protocol emphasizes Native-Source Data Curation. We aggregated independent datasets for Chinese and English contexts—spanning exams, guidelines, and case reports—to explicitly evaluate the LLM's mastery of region-specific medical norms and distinct clinical practice guidelines. This design avoids translation artifacts and ensures the Agent is tested against the specific regulatory and cultural frameworks of each healthcare system. During deployment simulations, we enforced a Language-Concordant Protocol: the LLM was constrained to process inputs and generate reasoning chains strictly in the source language. This methodological separation enables a stratified analysis of the LLM's "Bilingual Clinical Competency," verifying its feasibility as a localized clinical assistant capable of adhering to the specific standards of care in diverse linguistic environments.

**External Knowledge Integration via RAG**

To mitigate the inherent limitations of parametric memory and enhance adherence to clinical protocols, we implemented a Retrieval-Augmented Generation (RAG) architecture as a core "External Tool" for the Agent. The external knowledge base was constructed using the semantic guideline chunks derived during the Data Construction phase, ensuring a closed-loop validation of authoritative sources. We evaluated the impact of this knowledge augmentation from knowledge-oriented evaluations (Guideline-based Open QA) to workflow-based simulations (real-world-case-based dialogue simulations). Specifically, within the multi-turn dialogue simulations, the RAG mechanism was triggered at each interaction turn, retrieving relevant clinical protocols to inform the LLM's immediate response. This granular integration allows us to examine whether real-time knowledge injection enhances the agent's active inquiry and history-taking capabilities, rather than merely refining the final diagnostic output. This comparative analysis aims to quantify the efficacy of external retrieval in grounding the Agent's decision-making process, specifically assessing whether dynamic access to guideline protocols can optimize performance in complex, multi-turn clinical workflows compared to the Agent's intrinsic reasoning capabilities alone.

## Agent Reliability and Stochasticity Profiling

Given the inherent non-determinism of generative agents, ensuring consistent performance is critical for clinical safety. We employed a Worst-at-k methodology to quantify the lower bound of the LLM's performance reliability and assess the risk of sporadic, low-quality outputs (hallucinations or reasoning failures). We constructed a stratified reliability probe set by randomly selecting 74 representative cases (2 per assessment criterion) from the broader repository. To capture the distribution of potential behaviors, the LLM was tasked with independently resolving each case 10 times, generating a spectrum of 10 distinct performance scores per scenario: {$s_1, s_2,…, s_{10}$}.

To rigorously stress-test the LLM's safety floor, we computed the Worst@k metric. For a given iteration count k(ranging from 1 to 10), we simulated a "worst-case" deployment scenario by randomly sampling k outcomes from the generated set and isolating the minimum score. The aggregate reliability metric is defined as:

$$Worst@k = \frac{1}{M}\sum_{j=1}^{M}[min_{s \in Sample_k(R_j)}(s)]$$

where M denotes the total test cases (M=74), $R_j$ represents the response set for the j-th case, and $Sample_k(R_j)$ indicates the subset of k scores sampled without replacement. By analyzing the performance degradation curves across increasing k values, we derived a quantitative measure of the LLM's stability, distinguishing between robust clinical reasoning and fragile, high-variance behavior.

## Results

### Establishment of the Evaluation Framework and Global Performance Overview

To empirically chart the capability boundaries of dental LLMs, we executed a three-stage research design as illustrated in Figure 1, progressing from raw data acquisition to structured evaluation. First, regarding Data Acquisition, we aggregated a comprehensive bilingual repository comprising 4,258 standardized exam items (2,192 English, 2,066 Chinese), 60 authoritative clinical practice guidelines, and 118 real-world case reports (eTable 1). Second, to establish a "Clinical Alignment" standard, a three-round Delphi Consensus process was conducted. This process crystallized the evaluation criteria into 37 validated metrics, stratified into 16 Safety-Binary guardrails and 21 Effectiveness-Quantitative indicators, and instantiated a Risk Stratification Framework (Weights 1–5) (eTable 2). Third, during Test Set Construction, a "Human-in-the-loop" annotation workflow mapped the raw repository to these consensus metrics. This yielded a structured evaluation suite consisting of 205 guideline-based queries (covering 443 checkpoint instances and 1,225 rubric points) for knowledge evaluation and destructured the case reports into 498 checkpoint instances and 1,318 rubric points for dynamic clinical simulation (eTable 3), both achieving 100% coverage of the 37 consensus indicators. Subsequently, in the LLM Response Generation phase, the workflow bifurcated: MCQs and Open-end QAs were directed to the Target LLM to elicit choices with reasoning, while Case Vignettes initiated a multi-turn dialogue where the Target LLM (acting as a Doctor) interacted with a Simulated Patient under restrictive prompts to formulate a treatment plan. Finally, strictly aligned with the Scoring Metrics, a dual-scoring mechanism was implemented to operationalize these metrics: a "zero-tolerance" binary score (0/1) for safety checkpoints, and a normalized weighted score (mapped to [0, 1]) for effectiveness checkpoints. These criteria scores were then synthesized into weighted averages to generate comprehensive analysis charts.

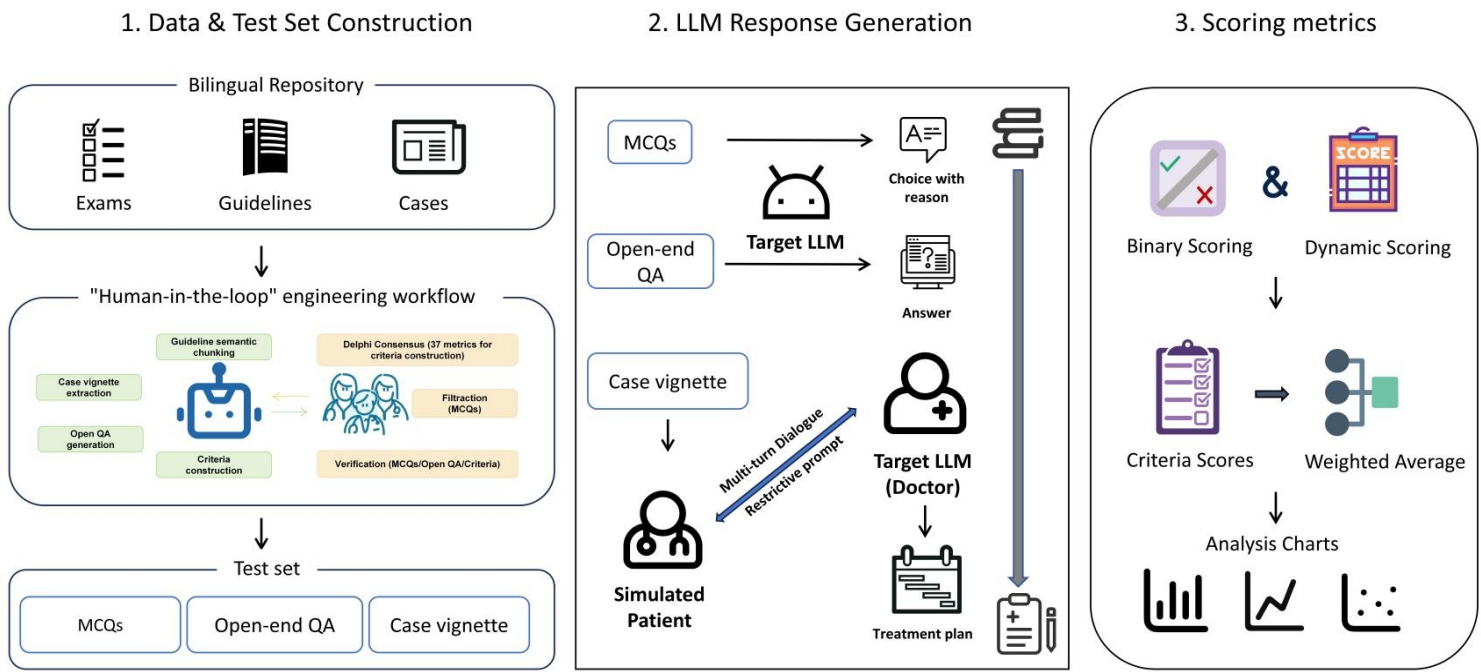

**Figure 1. Overview of the SCMPE Framework.** The figure and all its elements were designed and assembled using Microsoft PowerPoint graphics and publicly available vector icons obtained from https://www.svgrepo.com/, which provides free and open-license SVG resources that allow both academic and commercial use with modification.

To investigate the performance of various LLMs on this framework, we employed DeepSeek-V3.2-Exp (20251117), GPT-5.2 (GPT-5.2-2025-12-11), Claude-Sonnet-4.5 (20250929), Gemini-3-Pro (gemini-3-pro-preview-11-2025), Baichuan-M2 (Baichuan-M2-32B) and MedGPT (MedGPT-251130, Medlinker) as the test models. Notably, MedGPT serves as the representative vertical domain model in this study. We explicitly verified that its pre-training corpus did not include the specific MCQs, guidelines, or case reports constructed for this benchmark. This exclusion is critical to rule out data leakage, ensuring that the subsequent analysis of its performance trends reflects genuine domain generalization and reasoning capabilities rather than rote memorization. All evaluations were conducted within a comparable time window, specifically between November 2025 and December 2025. During the experiments, all parameters were kept at their default configurations.

We employed GPT-4.1 (gpt-4.1-2025-04-14) as the automated scoring engine. Figure 2 presents the global performance overview across the three resulting task types: MCQs, Open QA (Guideline), and Multi-turn Dialogue (Clinical Case). In MCQs, MedGPT achieved the highest Accuracy of 0.944 and Macro-F1 score with 0.855 (Figure 2a-b). Similarly, in Open QA (Guideline) (Figure 2c), models maintained competitive weighted normalized scores, with MedGPT (0.751) and GPT-5.2 (0.697) demonstrating proficiency in retrieving and synthesizing static protocols. However, the global statistics reveal a divergent trend in Multi-turn Dialogue (Clinical Case) (Figure 2d): despite the high theoretical baseline, weighted normalized total scores dropped notably in dynamic interactions. MedGPT—as a domain-specialized (vertical) medical model—still shows the same "Knowledge-Action Gap" degradation pattern when shifting from static knowledge tasks to dynamic clinical execution, although it remained among the top performers (0.503), followed closely by GPT-5.2 (0.494). Taken together, these results motivate a cross-task analysis to quantify how performance degrades when moving from MCQs and Open QA (Guideline) to Multi-turn Dialogue (Clinical Case), i.e., the Knowledge-Action Gap.

Notably, the generally low ROUGE-L scores observed in eFigure 1 (peaking at only ~0.314) highlight a critical methodological fallacy: evaluating "reasoning" capabilities through text overlap metrics is fundamentally flawed. Such text-matching assessments neither provide interpretability for the models' high accuracy in standardized exams (MCQs) nor possess any predictive validity for their ability to solve actual problems in complex clinical scenarios. This confirms a substantial disconnect between such abstract "reasoning evaluations" and true clinical competence, proving that these metrics fail to reflect the models' decision-making value in real-world medical environments.

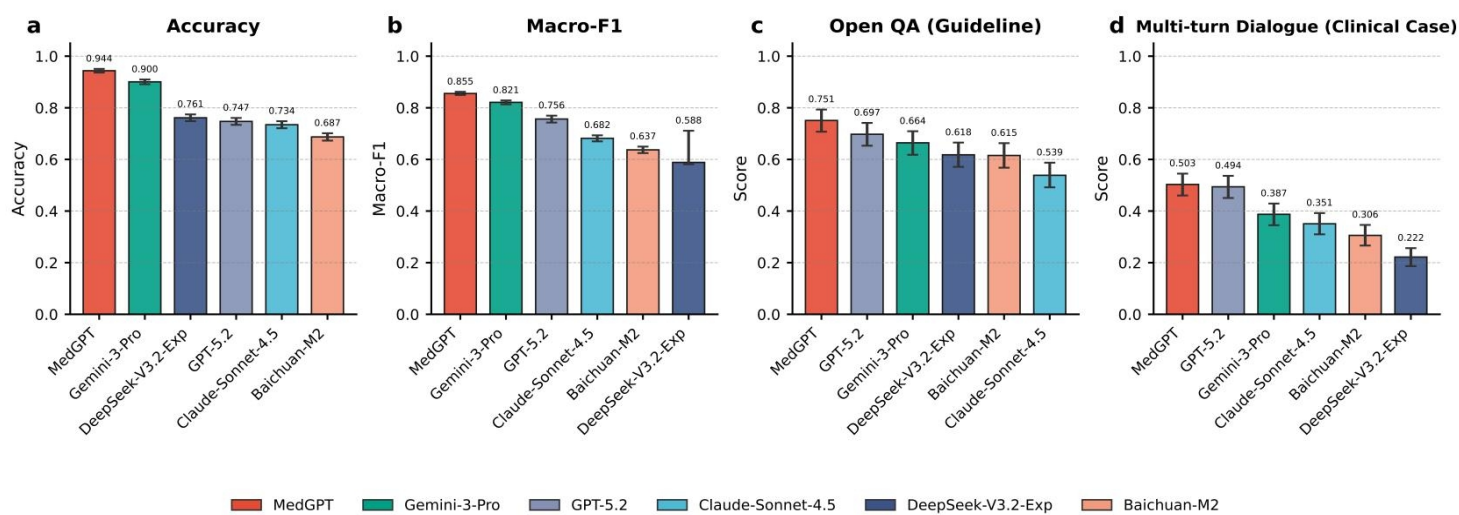

Figure 2. Comparative Evaluation of LLMs across MCQs, Open QA (Guideline), and Multi-turn Dialogue (Clinical Case). MedGPT refers to the MedGPT-251130 version. (a-b) Performance on the Standardized Exam (MCQ) task. Panel (a) displays the Accuracy scores. Panel (b) shows the Macro-F1 scores. (c) Performance on the Guideline-based Open QA task, measured by weighted normalized scores. (d) Performance on Clinical Case simulations (multi-turn dialogue). Models are ordered by their respective scores in each panel.

We assessed the agreement between model-generated scores and human expert ratings using standard concordance statistics. Overall, the alignment was strong, with Spearman ρ in [0.8286–0.9429] across experts, suggesting that the model-based scoring preserves both linear consistency and rank-order consistency with expert judgment. These results indicate that the model's scoring is well-calibrated to expert assessment and can serve as a reliable proxy for evaluating diagnostic reasoning quality.

Furthermore, to characterize agentic stochasticity and quantify a conservative "performance floor" for deployment, we conducted reliability profiling using the Worst@k metric (eFigure 2). Across k from 1 to 10, Worst@k scores decrease monotonically for all models, indicating that increasing the number of sampled responses exposes non-negligible low-end failures even when average performance appears competitive. The degradation pattern is task-dependent: models are comparatively more stable in MCQs, show a clearer drop in Guideline-based Open QA, and exhibit the steepest deterioration in Multi-turn Dialogue (Clinical Case) — suggesting that long-horizon interaction amplifies variance through missed information gathering, imperfect state tracking, and occasional reasoning breakdowns. Notably, top-tier models maintain a higher Worst@k trajectory across tasks, whereas weaker models show faster collapse as k increases, highlighting reliability—rather than point estimates alone—as a key differentiator for safety-critical use.

## Quantifying the Knowledge-Action Gap

Despite the high baseline competence in static tasks, a cross-task analysis revealed a precipitous decline in performance as models progressed from knowledge-oriented evaluations to workflow-based simulations. Figure 3a visualizes this "Knowledge-Action Gap" through a stratified performance comparison across languages. Regardless of whether the context is English or Chinese, average scores exhibit a consistent stepwise degradation from MCQs, to Open QA (Guideline), and finally to Multi-turn Dialogue (Clinical Case). This degradation is further quantified in the model-specific trajectory analysis (Figure 3b). Without exception, all evaluated models exhibited a steep negative slope when moving from single-turn selection tasks to multi-turn clinical management. The scores dropped significantly, with leading models falling from the 0.90–0.95 range in MCQs to approximately 0.50 in Multi-turn Dialogue. This trend empirically confirms the "high scores, limited capability" paradox: the ability to select the correct option in a structured exam does not linearly translate to the dynamic state-tracking and decision-making required in a clinical workflow. Importantly, this persistent decline across both general-purpose and domain-specialized models highlights the universality of the Knowledge-Action Gap. Moreover, performance in Multi-turn Dialogue suggests that current LLMs still operate primarily in a "Co-pilot / Assisted Mode" rather than being ready for unsupervised end-to-end clinical decision-making, leaving a substantial gap before independent clinical deployment is feasible.

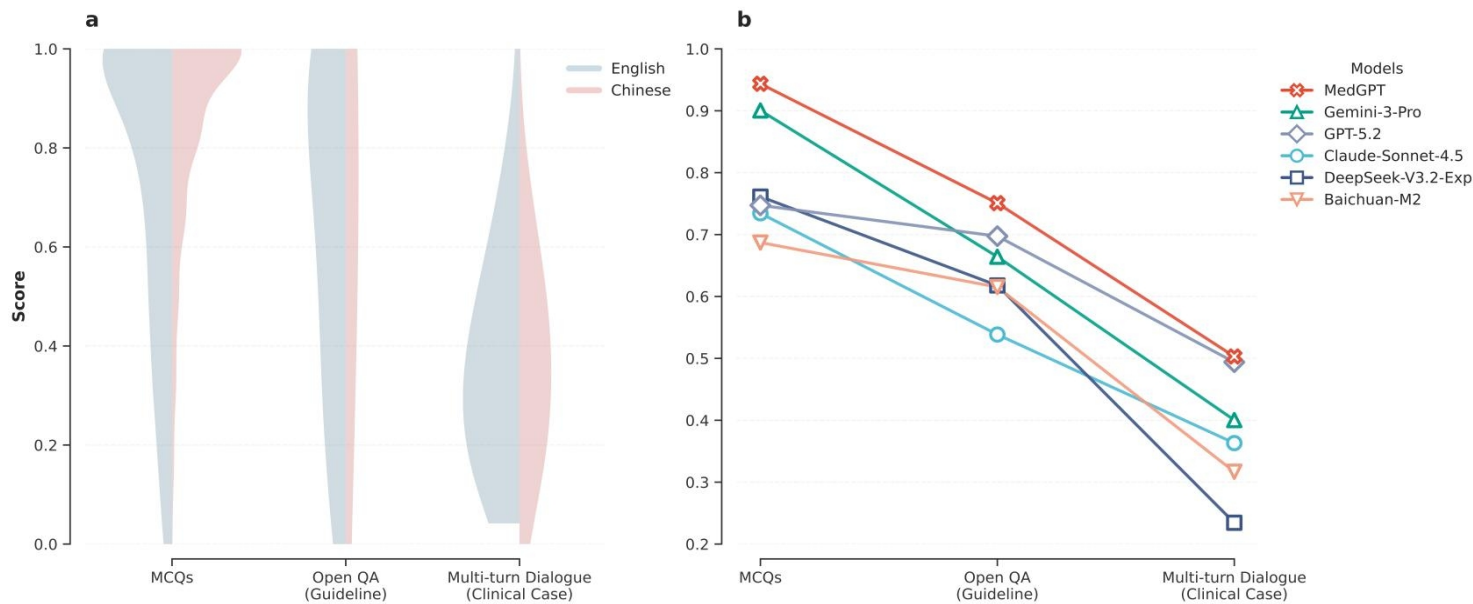

**Figure 3. The "Knowledge-Action Gap": Performance Degradation from Static Exams to Dynamic Clinical Simulations.** MedGPT refers to the MedGPT-251130 version. (a) Average performance comparison across three task types (MCQs, Open QA, and Multi-turn Dialogue) stratified by language (English vs. Chinese). A consistent decline in scores is observed as task complexity increases, regardless of the language. (b) Trajectory analysis of individual models across the three task types. All models display a steep negative slope, illustrating the sharp drop in performance when transitioning from standardized single-turn tasks to multi-turn clinical dialogue.

## Dialogue Efficiency and Information-Gathering Patterns in Multi-turn Dialogue

We adopted an interaction efficiency metric, the Efficiency Coefficient, to quantify information-gathering efficiency in multi-turn clinical dialogues. By jointly accounting for dialogue quality (Clinical Case score) and interaction cost (turn count), this metric directly captures score gain per turn in simulated history taking.

As shown in Figure 4, the top-performing models exhibit distinct high-performance strategies. MedGPT achieves strong Clinical Case scores with fewer turns, resulting in a higher Efficiency Coefficient, consistent with structured, goal-directed questioning that rapidly covers key history elements and converges to an actionable management plan. In contrast, GPT-5.2 also attains high Clinical Case scores but with more turns and a comparatively lower Efficiency Coefficient, suggesting a "thorough-accurate" strategy in which more granular information acquisition supports downstream decision quality; this may explain why a general-purpose model remains competitive in complex clinical dialogues. Overall, other models show lower efficiency distributions, specifically with DeepSeek-V3.2-Exp and Baichuan-M2 exhibiting notably fewer dialogue turn. This is mechanistically consistent with either premature closure due to incomplete history taking or diminished effective use of accumulated context in longer dialogues.

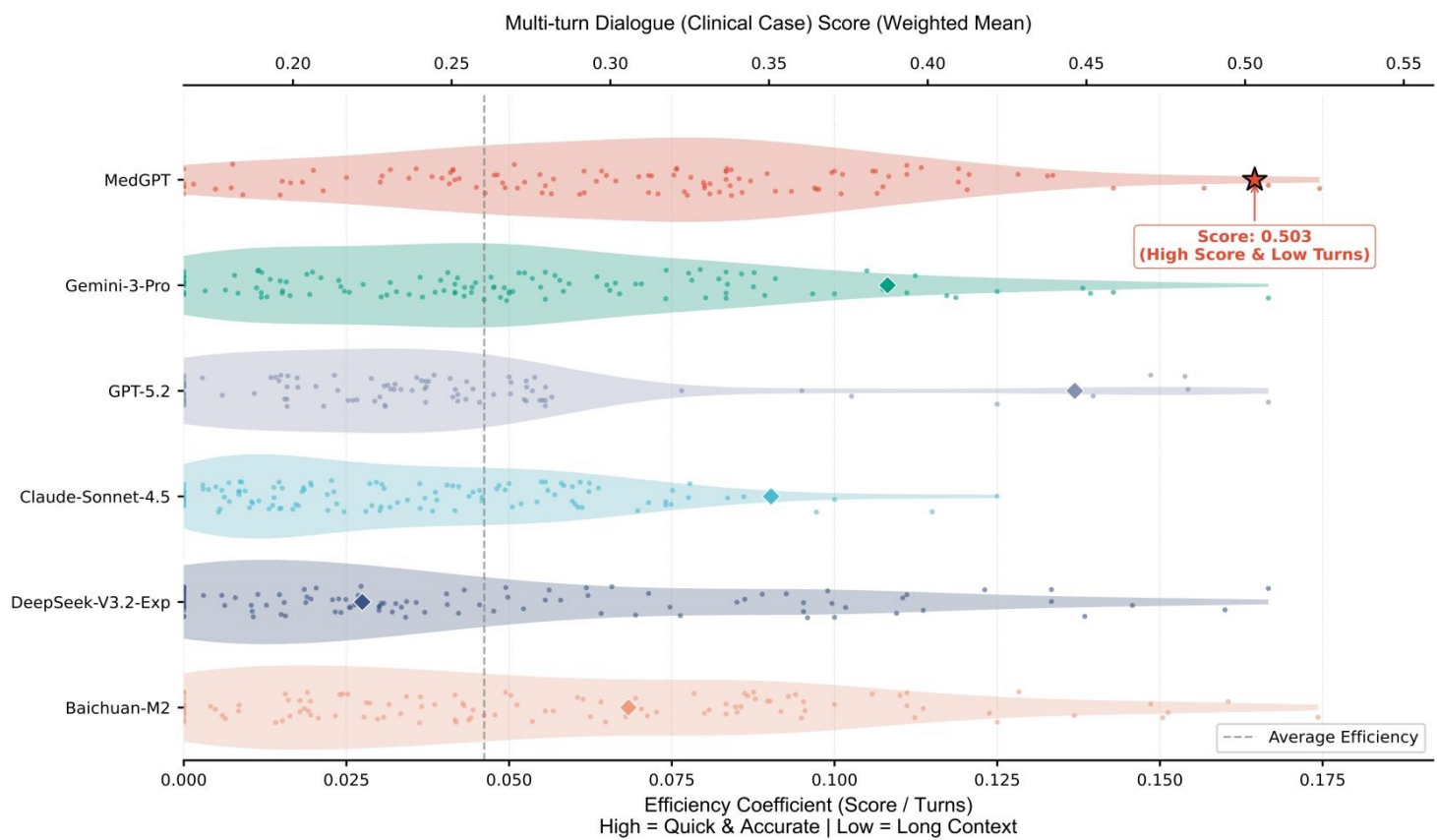

**Figure 4. Efficiency Coefficient and Performance Score in Multi-turn Clinical Dialogue.** MedGPT refers to the MedGPT-251130 version. The chart illustrates the trade-off between interaction cost and quality. The bottom x-axis represents the Efficiency Coefficient (calculated as Score / Turns), where the scatter points and violin plots depict the distribution of efficiency across individual simulation cases for each model. The top x-axis represents the Multi-turn Dialogue (Clinical Case) Score (Weighted Mean), which corresponds to the position of the central markers (stars and diamonds) for each model. MedGPT (Star) demonstrates an optimal balance with high scores and high efficiency, whereas other models show varying degrees of efficiency loss or premature closure.

To further localize the bottleneck in multi-turn performance, we conducted a control experiment designed to disentangle inquiry/action from reasoning/knowledge. In this setting, the full Case Vignette was directly provided to the target LLM, bypassing the inquiry phase and requiring immediate plan generation. Table 1 shows a consistent and substantial score increase for all models in the Direct Input Mode compared with the Multi-turn Inquiry Mode. This uniform uplift indicates that the primary limitation in multi-turn dialogues is not the availability of clinical knowledge or the ability to generate a coherent plan when information is complete, but rather deficits in proactive information gathering and dynamic state tracking during interaction; consequently, the Knowledge-Action Gap in this benchmark is driven predominantly by Information Gathering (Action) rather than Clinical Reasoning (Knowledge).

**Table 1.** Performance Comparison of LLMs in Multi-turn Inquiry Mode versus Direct Input Mode.

| LLMs | Multi-turn Inquiry Mode | Direct Input Mode |
|---|---|---|
| MedGPT | 0.5030 | 0.7468 |
| GPT-5.2 | 0.4939 | 0.7057 |
| Gemini-3-Pro | 0.3873 | 0.6617 |
| Claude-Sonnet-4.5 | 0.3506 | 0.6922 |
| Baichuan-M2 | 0.3057 | 0.6023 |
| DeepSeek-V3.2-Exp | 0.2218 | 0.5809 |

- All models show a statistically significant performance improvement in the Direct Input Mode compared to the Multi-turn Inquiry Mode ($p \leq 0.0001$). P-values are derived from weighted, bootstrap tests for all pairwise comparisons, adjusted using the Holm correction.
- MedGPT refers to the MedGPT-251130 version.

**Metric-level Disparities and Safety Vetoes in Dynamic Dialogues**

At the metric level, the radar plots in Figure 5 show broad coverage across Safety and Effectiveness checkpoints in static Guideline QA, whereas the polygons contract and distort in multi-turn Clinical Case simulations, indicating a systematic failure to translate static knowledge into dynamic execution. The dot-plot distributions in Figure 6 further demonstrate that this degradation is most pronounced on high-weight checkpoints and is frequently triggered by missing key questions during interaction. Because Safety is evaluated via a zero-tolerance binary veto, a single omission of safety-critical information can yield Safety=0 even when subsequent text appears plausible, thereby separating fluent-but-unsafe outputs from clinically acceptable behaviors.

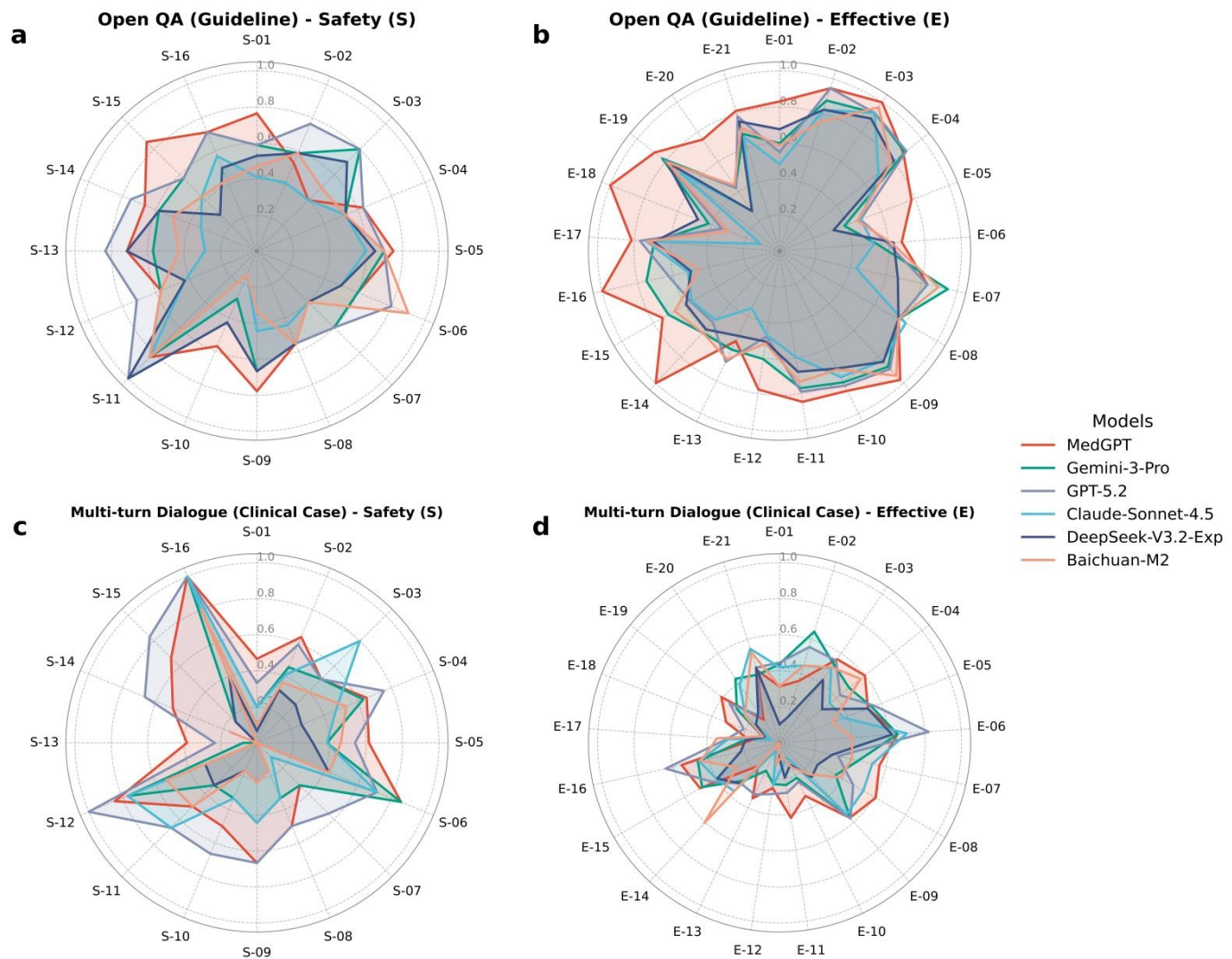

**Figure 5. Radar Chart Analysis of Consensus Metrics Coverage.** MedGPT refers to the MedGPT-251130 version. (a-b) Performance coverage in Guideline QA tasks for Safety (a) and Effectiveness (b) metrics. Note the broad, nearly complete polygons indicating strong theoretical retrieval. (c-d) Performance coverage in Clinical Case simulations. The significant shrinkage of the polygons illustrates the degradation in applying safety (c) and effectiveness (d) protocols in dynamic dialogue settings.

Representative low-score failure modes are observed in S-15 (Indications Assessment for Dental Implant Surgery, Weight 5) and S-13 (Prevention of Complications in Root Canal Therapy, Weight 2). In S-15, models often fail to systematically assess bone volume and anatomical constraints during the dialogue, preventing appropriate risk stratification and triggering a safety veto. In S-13, models may omit critical preoperative imaging assessments or procedural precautions, resulting in Safety=0 despite possessing theoretical knowledge. These patterns demonstrate that the scoring framework can surface potentially fatal decision errors that arise specifically from information gaps in multi-turn inquiry.

In contrast, several checkpoints remain comparatively stable in dynamic dialogues, suggesting safer zones for clinical copilot use. S-12 (Identification of Contraindications for Tooth Extraction, Weight 5) and S-11 (Recognition and Emergency Management of Anaphylactic Shock, Weight 4) more consistently follow standardized pathways, supporting more robust safety performance. Similarly, E-06 (Early Clinical Indicators of Oral Neoplasms, Weight 4) is often supported by distinct red-flag recognition (e.g., non-healing ulcers), yielding more reliable effectiveness in multi-turn settings. Collectively, Figure 5 and Figure 6 identify a mechanism by which specific high-complexity checkpoints are disproportionately vulnerable to omission-driven failures, while also delineating a subset of metrics that appear more reliable for risk-stratified human–AI collaboration.

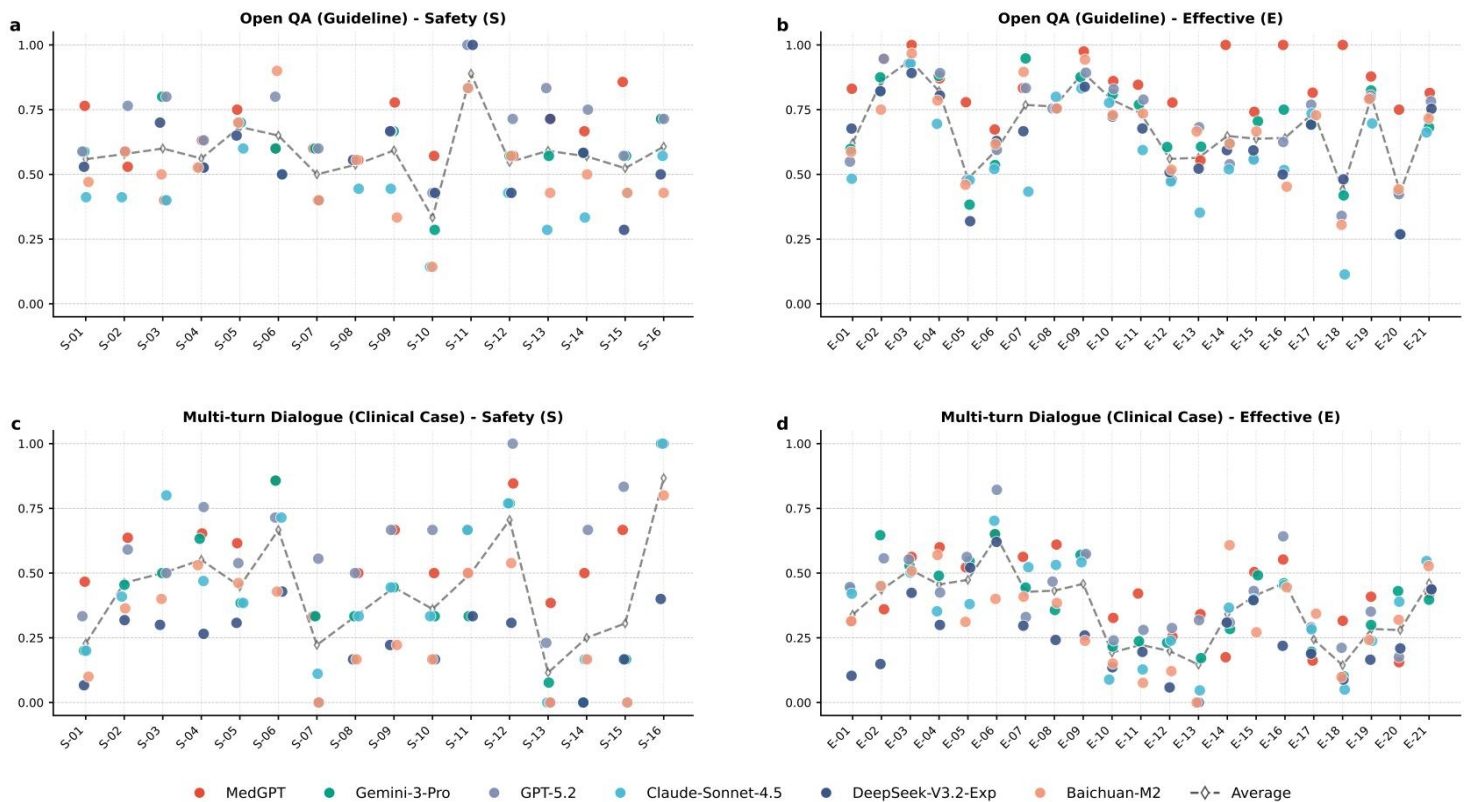

**Figure 6. Metric-Level Performance Discrepancy Analysis.** MedGPT refers to the MedGPT-251130 version. (a-b) Dot plots showing the distribution of model scores for each metric in Guideline QA. Models consistently score high on static knowledge metrics (e.g., S-11, E-03). (c-d) Dot plots for Clinical Case simulations. The "Average" dot represents the mean score across all models per metric.

## Mapping the Safe-Operating Boundaries and Clinical Trade-offs

To rigorously delineate the operational boundaries of autonomous dental agents, we mapped the capability space of LLMs regarding "Safety Guardrails" versus "Clinical Effectiveness," visualizing the performance trajectory from static Open QA (circles) to dynamic multi-turn simulations (triangles). This dual-axis analysis reveals a critical performance degradation, demonstrating that a model's ability to retrieve knowledge in static tasks does not inherently guarantee execution quality in dynamic workflows. As illustrated in Figure 7, all evaluated models retreat from the "Safe-Operating Region" (upper-right quadrant) observed in Open QA tasks. In the dynamic setting, models often exhibit a "Safety-Biased, Low Efficacy" profile, where the ability to synthesize comprehensive treatment plans (Effectiveness) collapses more precipitously than adherence to safety protocols. Notably, only select models such as GPT-5.2 and MedGPT manage to maintain a relatively robust safety floor (Safety > 0.6), separating themselves from the lower-left cluster. However, even these leading models display a substantial "Knowledge-Action Gap," characterized by a sharp decline in Effectiveness scores compared to their static baselines. This distribution underscores that the primary bottleneck in current architectures is not the trade-off of safety for efficacy, but rather the inability to maintain active state tracking and information gathering required to sustain high clinical effectiveness in multi-turn interactions.

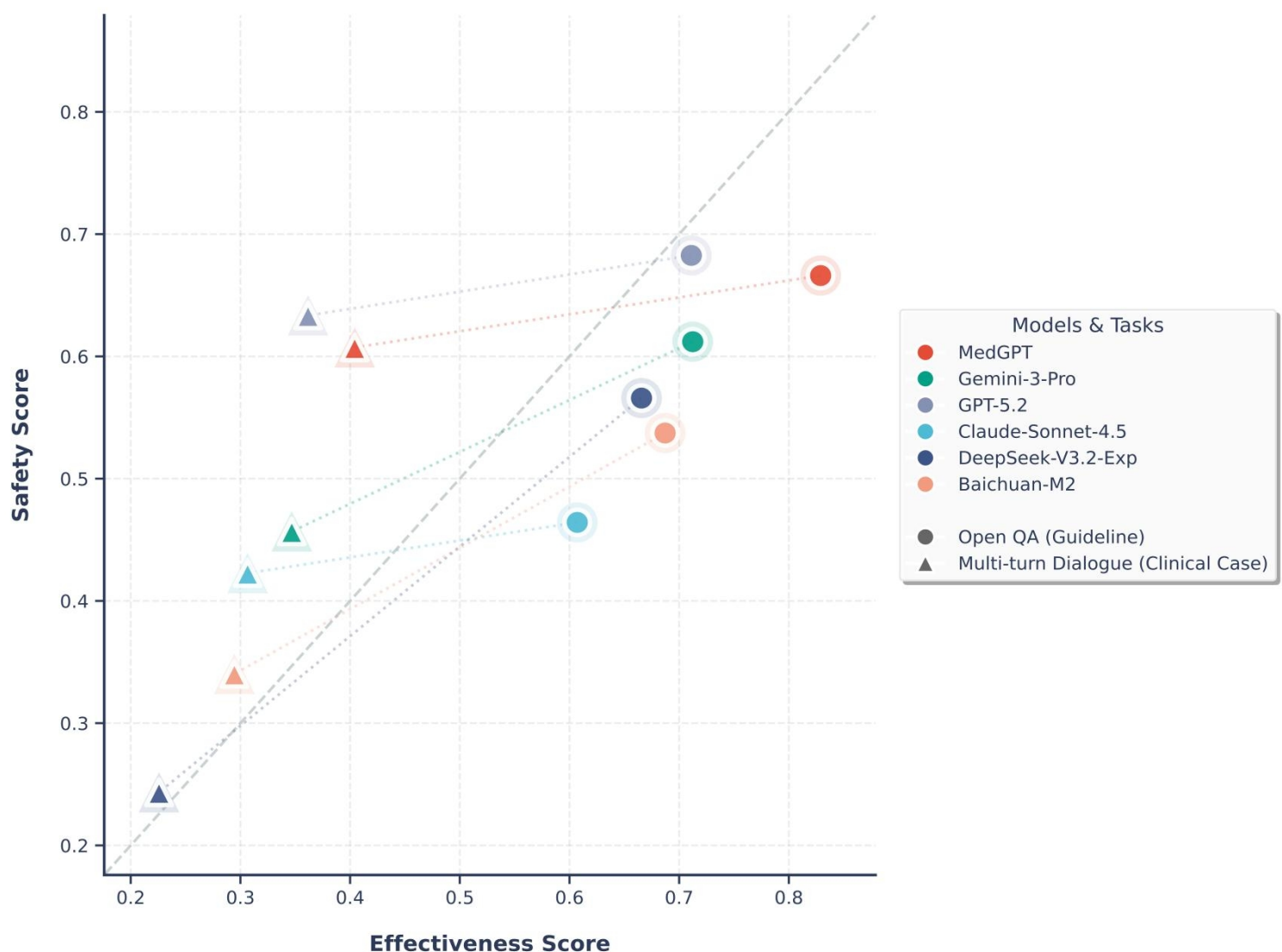

**Figure 7. Mapping the Knowledge-Action Gap: Safety vs. Effectiveness Trade-off.** MedGPT refers to the MedGPT-251130 version. The scatter plot visualizes the performance trajectory of Large Language Models from static knowledge tasks to dynamic clinical simulations. The X-axis represents the Clinical Effectiveness Score, and the Y-axis represents the Safety Score. Distinct markers indicate task types: circles represent Open QA (Guideline) and triangles represent Multi-turn Dialogue (Clinical Case), with dashed lines illustrating the performance shift for each model. The upper-right quadrant denotes the ideal "Safe-Operating Region." While models generally inhabit this high-performance zone in static QA tasks, a significant regression is observed in dynamic scenarios.

**External Knowledge Integration and Agentic Reliability**

We further evaluated the impact of Retrieval-Augmented Generation (RAG) as a remediation strategy. As presented in Table2, the integration of an external knowledge base containing the 60 semantic guideline chunks resulted in statistically significant improvements in Guideline-based Open QA tasks, with performance gains ranging from 0.0125 to 0.1626 across evaluated models. Notably, general models with lower initial baselines, such as Claude-Sonnet-4.5 and Baichuan-M2, exhibited the most substantial benefits in this static retrieval setting (0.1626 and 0.1318, respectively). However, the benefit in the Case Dialogue tasks was markedly heterogeneous and generally more modest for top-performing models. While DeepSeek-V3.2-Exp achieved a significant boost (0.1391), other leading models like MedGPT and GPT-5.2 showed limited improvements (0.0413 and 0.0357), and Baichuan-M2 even experienced performance degradation (-0.0147). This divergence demonstrates that while RAG successfully mitigates specific factual hallucinations—pulling the "Safety" coordinates rightward in the capability map—it fails to consistently bridge the "Knowledge-Action Gap." The inability of RAG to universally improve, or in some cases even maintain, performance in dynamic workflows underscores that external knowledge injection cannot fully compensate for intrinsic deficits in reasoning logic or state tracking required for complex clinical simulations.

**Table 2.** Impact of Retrieval-Augmented Generation (RAG) on LLM performance across Open QA and Multi-turn Dialogue tasks.

| LLMs | Mode | Open QA | Multi-turn Dialogue |
| --- | --- | --- | --- |
| MedGPT | No RAG | 0.7508 | 0.5030 |
|  | RAG | 0.7633 (+0.0125) | 0.5443 (+0.0413) |
| GPT-5.2 | No RAG | 0.6974 | 0.4939 |
|  | RAG | 0.7473 (+0.0499) | 0.5296 (+0.0357) |
| Gemini-3-Pro | No RAG | 0.6642 | 0.3873 |
|  | RAG | 0.6971 (+0.0329) | 0.4687 (+0.0814) |
| DeepSeek-V3.2-Exp | No RAG | 0.6178 | 0.2218 |
|  | RAG | 0.6770 (+0.0592) | 0.3609 (+0.1391) |
| Baichuan-M2 | No RAG | 0.6152 | 0.3056 |
|  | RAG | 0.7470 (+0.1318) | 0.2909 (-0.0147) |
| Claude-Sonnet-4.5 | No RAG | 0.5385 | 0.3506 |
|  | RAG | 0.7011 (+0.1626) | 0.3829 (+0.0323) |

**Impact of Language Concordance on Performance**

To evaluate the holistic robustness of LLMs across diverse medical environments, we analyzed performance on our natively sourced bilingual datasets (eFigure 3). Critically, this stratification extends beyond mere translation proficiency; since the Chinese and English samples were independently collected, they represent distinct medical-cultural contexts and guideline ecosystems. The results revealed a divergent pattern based on model architecture. While top-tier global generalist models (e.g., GPT-5.2, Gemini-3-Pro) exhibited a relatively balanced proficiency across languages, domain-specific and regional models (e.g., MedGPT, DeepSeek) displayed a discernible "Native-Language Alignment." Specifically, MedGPT achieved peak performance in its primary linguistic context (Chinese) but experienced a notable decline in the non-native setting (English), particularly in Multi-turn Dialogue (dropping from ~0.58 to ~0.41). Similarly, regional models like DeepSeek showed significant disparities in MCQs. This suggests that while general clinical reasoning capabilities are transferable, the deep cultural and guideline-specific nuances in dentistry still impose a dependency on the pre-training distribution, challenging the notion of universal robustness for specialized agents. Thus, true clinical robustness requires not just reasoning capability, but explicit alignment with the heterogeneous medical standards of diverse linguistic ecosystems.

# Discussion

Our study proposes a paradigm shift in evaluating LLMs in dentistry, moving beyond the traditional "Knowledge Engine" perspective to a comprehensive "Clinical Agent" framework. The observed performance drop from knowledge-oriented evaluations to workflow-based simulations offers critical insights into the architectural requirements of autonomous dental agents. Conceptually, our evaluation framework probes the same underlying agentic competencies under two task regimes—from knowledge-oriented evaluations to workflow-based simulations—thereby stress-testing not only recall of protocols but also interactive planning, state tracking, and decision safety. While recent studies confirm that LLMs can achieve passing scores in dental licensing examinations[9, 10], our results demonstrate that high proficiency in standardized exams (MCQs) does not guarantee the successful execution of complex, multi-step clinical workflows. We attribute the precipitous performance decline observed in our results to the escalation of "Task Complexity" and the challenge of "Information Asymmetry," rather than merely an increase in question difficulty. Unlike static exams where all conditions are provided, clinical scenarios require the Agent to actively bridge the information gap through strategic inquiry in an incomplete information setting. This finding corroborates the "High scores vs. Limited capability" paradox and highlights the critical challenge of the "Faithfulness-Plausibility Gap," where models may generate plausible outputs without faithful underlying reasoning logic[22]. Future development must prioritize the integration of Tool Use capabilities—specifically, the ability to actively query and synthesize external guidelines—rather than merely scaling parameter size for rote memorization[11].

The introduction of the Simulated Patient (SP) multi-turn dialogue assessment establishes a higher standard of ecological validity compared to single-turn QA benchmarks. In our end-to-end simulations, the LLM is required not only to diagnose but to maintain State Tracking across the entire clinical trajectory—from history taking to treatment planning and follow-up. This distinction is vital for differentiating a conversational "chatbot" from a task-oriented "Clinical Agent" capable of handling long-horizon tasks[23]. The performance drop observed when moving from knowledge-oriented evaluations to workflow-based simulations, particularly in long-context scenarios, highlights the fragility of current LLMs in maintaining coherent clinical logic over time. Furthermore, by utilizing Native-Source Data rather than translated datasets, we further substantiate that these performance deficits are intrinsic to the model's active inquiry and strategic planning capabilities rather than artifacts of linguistic misalignment.

The integration of RAG transforms the LLM's operation from opaque generation to a transparent, evidence-based process. By anchoring responses to specific guideline chunks, the Traceable RAG mechanism significantly mitigates the hallucination of non-existent treatment protocols. However, our results reveal that this benefit does not uniformly translate to workflow-based simulations. The observation that RAG yielded negligible gains—or even performance regression in models with weaker reasoning cores (e.g., Baichuan-M2)—indicates that simply accessing external guidelines is insufficient. It suggests that without a stable "Decision Core" to filter and utilize retrieved information, external knowledge can become cognitive noise rather than aid. Thus, external knowledge integration requires a stronger reasoning or Agentic framework to effectively solve clinical problems. Furthermore, current implementations reveal structural limitations: the retrieval component often relies on English-centric embedding models, while the generation component predominantly utilizes generic LLMs rather than medically optimized ones, with applications largely restricted to QA, report generation, and summarization. Crucially, existing evaluations disproportionately focus on accuracy and fluency, leaving bias and safety considerations severely underaddressed. Medical RAG remains in its nascent stage, necessitating future efforts to strengthen validation on private data, enhance multilingual capabilities, and establish stricter ethical and safety evaluation standards[24]. Access to external knowledge alone is insufficient without robust reasoning engines to interpret that knowledge within a specific patient context[25, 26].

The consistent superiority of MedGPT across the full evaluation continuum provides compelling evidence for the necessity of domain-adaptive pre-training in constructing autonomous Clinical Agents. This finding aligns with recent observations that general medical models perform suboptimally in the highly specialized field of dentistry due to a lack of deep domain knowledge. Consequently, efforts such as DentalBench highlight the critical importance of constructing sub-field specific benchmarks and corpora (e.g., for dentistry) for developing trustworthy medical AI[27]. MedGPT's dominance suggests that medical domain specificity extends beyond mere vocabulary retention to the internalization of "clinical syntax"—the implicit logical structures governing diagnosis and treatment planning. In the context of our agentic evaluation framework, this finding has profound implications for the Architectural Alignment of dental agents. A domain-specific model does not simply act as a larger database; it functions as a more stable Decision Core, capable of executing State Tracking in multi-turn dialogue simulations with higher fidelity[28].

Our study acknowledges several methodological constraints inherent to the evaluation framework. The composition and size of the expert panel may limit the generalizability of the consensus-based metrics, potentially introducing selection bias regarding specific clinical sub-specialties. Furthermore, the structured feedback mechanism employed during the Delphi process could introduce conformity or anchoring bias, leading to an artificial convergence of expert opinions. Regarding the inevitable subjectivity in human evaluation of long-context dialogues, we defend our strategy of employing "fine-grained checkpoint mapping + automated scoring" to maximize "Machine Consistency." Nevertheless, as LLM tasks evolve into open-ended, complex Agentic workflows, human evaluation becomes prohibitively expensive and hard to scale, while traditional single "LLM-as-a-judge" approaches often suffer from bias and limited perspective. In contrast, the "Agent-as-a-Judge" paradigm—capable of using tools, maintaining memory, and evaluating intermediate steps like a human—represents a critical path for scalable, low-cost, and nuanced assessment[29]. Although the Agent-as-a-Judge paradigm cannot fully replace human supervision, it serves as a necessary evolution for evaluating next-generation clinical agents. Future work must incorporate real-world evidence (RWE) and multicenter validation to further optimize these thresholds and conduct rigorous sensitivity analyses to ensure the robustness of the Safety Checkpoints across diverse clinical settings[30].

Crucially, while this study explores the feasibility of autonomous agents, the evaluation is strictly confined to the Dynamic Reasoning Core—effectively validating the "cognitive brain" of the agent rather than a fully functional "embodied agent." The current framework assesses text-based decision-making but does not evaluate the Action Execution capabilities required for physical intervention or direct integration with Electronic Health Record (EHR) systems. Consequently, the "autonomy" demonstrated here represents a potential for clinical reasoning rather than a readiness for physical clinical practice.

Although our Simulated Patient scenarios are derived from real-world reports, they inherently undergo a degree of structuring and sanitization during the prompt engineering process. In actual clinical practice, agents must contend with unstructured, high-noise inputs, including vague patient descriptions, colloquialisms, and ambiguous intent. Our results, therefore, may underestimate the difficulty of State Tracking in raw, noisy clinical environments. The robustness of these models against non-standard, noisy patient narratives remains a critical frontier. Future benchmarks must evolve from structured simulations to "in-the-wild" stress tests to fully verify the agent's resilience against the unpredictability of real-world patient interactions[31].

The SCMPE benchmark proposed in this study not only reveals the "Knowledge-Action Gap" but also provides an empirical blueprint for transitioning from mere LLM benchmarking to practical Clinical Agent deployment. Between 2024 and 2025, related research has grown exponentially, marking a technological shift from single models to systems capable of autonomy, planning, and tool use. However, Medical Agents differ significantly from general-purpose agents; they must operate under strict safety, privacy regulations, and ethical constraints, characterized by a "zero-tolerance" high-risk profile[32].

In this context, AI-driven medical software projects represent not just a technological upgrade but a reshaping of clinical workflows. The core challenge lies in balancing agent autonomy with controllability and trust. Driven by AI algorithm models, these projects achieve diagnostic support through data integration, logical computation, and decision assistance, with their core value lying in the balance between algorithm precision and clinical adaptability. Based on our findings, we delineate the application scenarios as follows:

- Doctor Side (Decision Support): The Agent can function as a "clinical copilot," assisting physicians in comprehensive history taking to improve efficiency, and performing real-time guideline checks to prevent omissions, thereby enhancing diagnostic efficiency and optimizing judgment accuracy.
- Patient Side (Health Education): In direct-to-patient scenarios, the Agent's function must be strictly limited to health education and consultation, explicitly prohibiting autonomous diagnosis or treatment to maintain safety boundaries.
- Checkpoint Side (Human-AI Collaboration): Based on our scoring mapping, clinicians can place higher trust in "high-scoring checkpoints," whereas "low-scoring checkpoints" involving complex reasoning require a "Human-in-the-loop" mechanism for manual review and intervention.

This stratified strategy aims to extend the boundaries of medical services while ensuring the safety of clinical practice by clearly defining service limits.

In conclusion, this study marks a pivotal step in redefining the evaluation of Large Language Models in dentistry, transitioning from the static paradigm of a Knowledge Engine to the dynamic, functional assessment of a Clinical Agent. The results reveals a critical dichotomy: while current models demonstrate impressive proficiency in the Static Knowledge Module, their capability within the Dynamic Reasoning Core remains fragile. Crucially, our findings indicate that simply accessing external guidelines via RAG is insufficient to bridge the "Knowledge-Action Gap" in complex state tracking; instead, domain-adaptive pre-training proves essential for constructing a stable "Decision Core" capable of adhering to clinical syntax. The paradox of "High scores vs. Limited capability" underscores that passing a board exam is a necessary but insufficient condition for autonomous practice, and safety guardrails must be enforced as zero-tolerance constraints rather than trade-offs for efficacy. Ultimately, this work provides not just a benchmark, but a foundational roadmap for aligning the "silicon reasoning" of AI with the "biological safety" required in patient care, establishing the rigorous boundaries within which Dental Agents can safely evolve. Only through such rigorous validation can we ensure that the promise of AI-driven patient empowerment—providing accessible, high-quality guidance for shared decision-making—is realized safely and effectively.

## Data availability

## Code availability

## Acknowledgements

We would also like to thank all the physicians for their contributions. This work would not have been possible without the insight and generosity of the physicians who contributed their time and expertise to SCMPE Benchmark.

## Author contributions

## Conflict of interest

## Funding

Funding for this study was supported by Beijing Chaoyang Digital Health Proof of Concept Center No.2025SLZZ030 and the National Natural Science Foundation of China for Young Scientists (82401196).

## Ethics Statement

All data sources we use to construct the SCMPE benchmark dataset are publicly available and free to use without copyright infringement. All questions in the SCMPE dataset have been appropriately anonymized so that they do not contain sensitive private information about patients. We do not foresee any other possible negative societal impacts of this work.

## References

1. Meng, Z. *et al*. DentVLM: A multimodal vision-language model for comprehensive dental diagnosis and enhanced clinical practice. arXiv:2509.23344 (2025).

2. Bedi, S. *et al.* MedHELM: Holistic evaluation of large language models for medical tasks. arXiv:2505.23802 (2025).

3. Farhadi Nia, M., Ahmadi, M. & Irankhah, E. Transforming dental diagnostics with artificial intelligence: advanced integration of

ChatGPT and large language models for patient care. *Front Dent Med* **5**, 1456208, doi:10.3389/fdmed.2024.1456208 (2024).

4. Jin, D. *et al.* What disease does this patient have? a large-scale open domain question answering dataset from medical exams. *Applied Sciences* **11**, 6421 (2021).

5. Pal, A., Umapathi, L. K. & Sankarasubbu, M. MedMCQA: A large-scale multi-subject multi-choice dataset for medical domain question answering. In Conference on Health, Inference, and Learning 248-260 (PMLR, 2022).

6. Jin, Q. *et al.* Pubmedqa: A dataset for biomedical research question answering. In Proceedings of the 2019 Conference on Empirical Methods in Natural Language Processing and the 9th International Joint Conference on Natural Language Processing (EMNLP-IJCNLP) 2567-2577 (2019).

7. Singhal, K. *et al.* Large language models encode clinical knowledge. *Nature* **620**, 172-180 (2023).

8. Gong, E. J., Bang, C. S., Lee, J. J. & Baik, G. H. Knowledge-practice performance gap in clinical large language models: systematic review of 39 benchmarks. *J Med Internet Res* **27**, e84120, doi:10.2196/84120 (2025).

9. Liu, M. *et al*. Large language models in dental licensing examinations: systematic review and meta-analysis. Int Dent J **75**, 213-222, doi:10.1016/j.identj.2024.10.014 (2025).

10. Sabri, H. *et al.* Performance of three artificial intelligence (AI)-based large language models in standardized testing; implications for AI-assisted dental education. J Periodontal Res **60**, 121-133, doi:10.1111/jre.13323 (2025).

11. Kim, H.-S. & Kim, G.-T. Can a large language model create acceptable dental board-style examination questions? A cross-sectional prospective study. J Dent Sci **20**, 895-900, doi:10.1016/j.jds.2024.08.020 (2025).

12. Nguyen, H. C., Dang, H. P., Nguyen, T. L., Hoang, V. & Nguyen, V. A. Accuracy of latest large language models in answering multiple choice questions in dentistry: A comparative study. PLoS One **20**, e0317423, doi:10.1371/journal.pone.0317423 (2025).

13. Ozdemir, Z. M. & Yapici, E. Evaluating the accuracy, reliability, consistency, and readability of different large language models in restorative dentistry. J Esthet Restor Dent **37**, 1740-1752, doi:10.1111/jerd.13447 (2025).

14. Thaliyil, K. P. *et al.* Evaluating the evidence-based potential of six large language models in paediatric dentistry: a comparative study on generative artificial intelligence. Eur Arch Paediatr Dent, doi:10.1007/s40368-025-01012-x (2025).

15. Johri, S. et al. An evaluation framework for clinical use of large language models in patient interaction tasks. Nat Med **31**, 77-86, doi:10.1038/s41591-024-03328-5 (2025).

16. Rewthamrongsris, P., Burapacheep, J., Trachoo, V. & Porntaveetus, T. Accuracy of large language models for infective endocarditis prophylaxis in dental procedures. Int Dent J **75**, 206-212, doi:10.1016/j.identj.2024.09.033 (2025).

17. Wu, X. *et al.* A multi-dimensional performance evaluation of large language models in dental implantology: comparison of ChatGPT, DeepSeek, Grok, Gemini and Qwen across diverse clinical scenarios. BMC Oral Health **25**, 1272, doi:10.1186/s12903-025-06619-6 (2025).

18. Liu, J. *et al.* Benchmarking large language models on cmexam-a comprehensive chinese medical exam dataset. *Advances in Neural Information Processing Systems* **36**, 52430-52452 (2023).

19. Wang, X. *et al.* Cmb: A comprehensive medical benchmark in chinese. In Proceedings of the 2024 Conference of the North American Chapter of the Association for Computational Linguistics: Human Language Technologies (Volume 1: Long Papers) 6184-6205 (2024).

20. Umapathi, L. K., Pal, A. & Sankarasubbu, M. Med-HALT: Medical Domain Hallucination Test for Large Language Models. arXiv:2307.15343 (2023).

21. Fleming, S. L. *et al.* MedAlign: A Clinician-Generated Dataset for Instruction Following with Electronic Medical Records.

In Thirty-Eighth AAAI Conference on Artificial Intelligence, doi:10.1609/AAAI.V38I20.30205 (2024).

22. Wang, W. *et al*. Medical reasoning in the era of LLMs: a systematic review of enhancement techniques and applications. arXiv:2508.00669 (2025).

23. Singh, S. & Beniwal, H. A survey on near-human conversational agents. Journal of King Saud University - *Computer and Information Sciences* **34**, 8852-8866, doi:10.1016/j.jksuci.2021.10.013 (2022).

24. Yang, R. *et al.* Retrieval-augmented generation in medicine: a scoping review of technical implementations, clinical applications, and ethical considerations. arXiv:2511.05901 (2025).

25. Pingua, B. *et al.* Medical LLMs: Fine-Tuning vs. Retrieval-Augmented Generation. *Bioengineering (Basel)* **12**, 687, doi:10.3390/bioengineering12070687 (2025).

26. Teo, Z. L. *et al*. Generative Artificial Intelligence in Medicine. *Nat Med* **31**, 3270-3282, doi:10.1038/s41591-025-03983-2 (2025).

27. Zhu, H., Xu, Y., Li, Y., Meng, Z. & Liu, Z. DentalBench: benchmarking and advancing LLMs capability for bilingual dentistry understanding. arXiv:2508.20416 (2025).

28. Chen, A. *et al.* Generalists vs. Specialists: Evaluating LLMs on Highly-Constrained Biophysical Sequence Optimization Tasks. *In Proceedings of the 42nd International Conference on Machine Learning* (2025).

29. Yu, F. When AIs judge AIs: the rise of agent-as-a-judge evaluation for LLMs. arXiv:2508.02994 (2025).

30. Kolbinger, F. R. & Kather, J. N. Adaptive validation strategies for real-world clinical artificial intelligence. *Nat Comput Sci* **5**, 980-986, doi:10.1038/s43588-025-00901-x (2025).

31. Javed, H., El-Sappagh, S. & Abuhmed, T. Robustness in deep learning models for medical diagnostics: security and adversarial challenges towards robust AI applications. *Artif Intell Rev* **58**, 12, doi:10.1007/s10462-024-11005-9 (2025).

32. Xu, G. *et al*. A comprehensive survey of AI agents in healthcare. TechRxiv (2025).

# Supplementary information

eTable 1. Data collection of the Bilingual Repository

| Dataset type  | Chinese | English |
|---------------|---------|---------|
| MCQs          | 2,066   | 2,192   |
| Guidelines    | 19      | 41      |
| Clinical cases| 68      | 50      |

eTable 2. Three-round scoring and final weights for Safety and Effectiveness metrics.

| ID | Type | Metrics Name | R1_med/IQR | R2_med/IQR | R3_med/IQR | Weight | clinical_description |
|---|---|---|---|---|---|---|---|
| S-01 | Safety | Selection, Dosage, and Contraindications of Local Anesthetics in Dentistry | 3.5/1.75 | 4/0 | - | 4 | Local anesthetics in dentistry are used to ensure patient comfort during invasive or potentially painful procedures such as restorative treatment, tooth extraction, endodontic therapy, and periodontal surgery. Specific scenarios include tooth preparation, exodontia, management of dental trauma, and debridement with suturing. |
| S-02 | Safety | Rational Use of Antibiotics in Dental Practice | 3/0.75 | - | - | 2 | Antibiotic use aims to achieve precise pathogen coverage, reduce antimicrobial resistance, and minimize adverse reactions. Clinical scenarios include complex extractions, implant surgery, and preprocedural prophylaxis for patients at high risk of infective endocarditis, as well as therapeutic management of acute alveolar abscess, osteomyelitis of the jaws, fascial space infections, necrotizing ulcerative gingivitis, suppurative sialadenitis, and periodontitis. |
| S-03 | Safety | Drug–Drug Interaction Alerts in Dental Pharmacotherapy | 4/3 | 4/1.75 | 4/0 | 4 | Polypharmacy is common in dental care, necessitating vigilance for drug–drug interactions that may diminish efficacy, increase toxicity, or raise adverse event risk. Concomitant use of epinephrine-containing local anesthetics with cardiovascular medications may precipitate cardiovascular events; combining antibiotics with anticoagulants increases bleeding risk; co-administration of opioid analgesics with antidepressants may lead to central nervous system depression. |
| S-04 | Safety | Safety Assessment for Special Populations (Pediatric, Pregnant/Postpartum, Elderly, and Polypharmacy) | 4/0.75 | - | - | 4 | Pediatric patients require attention to local anesthetic toxicity, airway management, and sedation risks; pregnant and peripartum patients warrant caution with radiographic exposure and medications such as tetracyclines that can affect the fetus; older adults require consideration of cardiovascular disease, hepatic/renal impairment, and cognitive disorders. |
| S-05 | Safety | Perioperative Management of Patients on Anticoagulation | 4/1 | - | - | 4 | Preoperative assessment should address thrombotic risk, bleeding risk, and the type of anticoagulant used; adjust preoperative anticoagulant dosing according to the procedure's bleeding risk; select appropriate hemostatic techniques postoperatively; and anticipate special risks for delayed or refractory postoperative hemorrhage. |
| S-06 | Safety | Risk Assessment of Infection Spread in Oral and Maxillofacial Regions | 4/1 | - | - | 4 | The risk of infection spread is influenced by anatomical location—e.g., floor-of-mouth infections are more likely to disseminate and cause emergencies—and by systemic status, with patients who have diabetes or chronic corticosteroid use being more susceptible to infection progression. |
| S-07 | Safety | Emergency Management of Temporomandibular Joint Dislocation | 3/0.75 | - | - | 2 | Excessive mouth opening, trauma, or structural abnormalities may cause condylar displacement from the glenoid fossa; rapid reduction is required to avoid complications (e.g., capsular injury, chronic pain). Reduction techniques should be selected based on etiology, with concurrent prevention of complications. |
| S-08 | Safety | Risk Assessment for Major Oral Hemorrhage | 4/1.5 | 4/0.75 | - | 4 | Risk stratification for oral hemorrhage should integrate etiology (trauma, postoperative bleeding, coagulopathy, neoplasm), bleeding rate, and systemic condition, followed by tiered interventions. Scenarios include secondary post-extraction hemorrhage, traumatic oral mucosal lacerations with arterial bleeding, and bleeding from rupture of advanced oral cancers. |
| S-09 | Safety | Emergency Management of Airway Obstruction in Dental Settings | 5/0.75 | - | - | 5 | Airway obstruction in dental settings is a life-threatening emergency, potentially caused by foreign body aspiration, postoperative tissue edema, allergic reactions, or sedative adverse effects. Rapid identification of obstruction type (partial vs. complete) and stratified intervention are essential to maintain oxygenation and prevent hypoxic brain injury or death. |
| S-10 | Safety | Early Signs Recognition of Local Anesthetic Systemic Toxicity (LAST) | 4/0.75 | - | - | 4 | Local anesthetic systemic toxicity (LAST) in dental care may result from intravascular injection, overdose, or rapid systemic absorption. Early signs can be subtle and progress quickly; prompt recognition and intervention are crucial to prevent life-threatening central nervous system or cardiovascular depression. Common early manifestations include CNS excitation, CNS depression, and cardiovascular depression. |
| S-11 | Safety | Recognition and Emergency Management of Anaphylactic Shock | 3.5/2.5 | 4/0.75 | - | 4 | Type I hypersensitivity reactions to drugs (e.g., local anesthetics, antibiotics) or materials (e.g., latex gloves) in dentistry present with acute circulatory collapse and multiorgan hypoxia and can rapidly progress to death. Early recognition and epinephrine-first management are key to successful resuscitation. |
| S-12 | Safety | Identification of Contraindications for Tooth Extraction | 5/0.75 | - | - | 5 | Pre-extraction assessment must exclude conditions that pose high intra- or postoperative risk (e.g., massive bleeding, infection spread, organ failure), including myocardial infarction within 6 months, unstable angina, uncontrolled hypertension, severe arrhythmias, hemophilia, long-term anticoagulant use, chronic immunosuppressive therapy, post-chemoradiation marrow suppression, extractions after bisphosphonate therapy, malignant tumor invasion, and acute infection. |
| S-13 | Safety | Prevention of Complications in Root Canal Therapy | 3/0.75 | - | - | 2 | Through precise preoperative imaging assessment, meticulous technique, and standardized postoperative care, risks such as instrument separation, canal perforation, and postoperative infection can be minimized. |
| S-14 | Safety | Risk Assessment in Orthodontic Treatment | 3.5/1 | - | - | 3 | Orthodontic risk assessment is a core pre-treatment step that integrates dentoalveolar features, systemic status, and patient adherence to prevent complications such as root resorption, periodontal damage, and temporomandibular disorders (TMD). Biomechanical risks include root resorption, periodontal tissue injury, and TMD; patient adaptability risks include poor compliance and interference with growth and development; special populations include patients with periodontitis and systemic diseases. |
| S-15 | Safety | Indications Assessment for Dental Implant Surgery | 4/1.5 | 4.5/1 | - | 5 | Indications for implant surgery should synthesize anatomical conditions (e.g., bone volume, proximity to neural structures), systemic health (e.g., diabetes, history of head-and-neck radiotherapy, smoking, periodontitis), and functional–esthetic demands (e.g., anterior esthetics) to avoid surgical failure and long-term complications. |
| S-16 | Safety | Systemic Risk Management in Periodontal Therapy | 3.5/1.75 | 3/0.75 | - | 2 | Systemic risk management in periodontal therapy aims to identify how systemic health (e.g., diabetes, cardiovascular disease) affects treatment, while preventing systemic complications related to infection spread, drug interactions (e.g., anticoagulants), or procedural provocation. |
| E-01 | Effectiveness | Guideline-Based Systematic Differential Diagnosis in Dentistry | 2.5/1 | - | - | 1 | Pain-based differential diagnosis should integrate characteristics, timing, and triggers to distinguish trigeminal neuralgia, pulpitis, apical periodontitis, temporomandibular disorders, and atypical facial pain; TMD etiologies should be differentiated based on clinical patterns; guideline-based prioritization should flag life-threatening conditions (e.g., malignant jaw tumors, rapidly spreading acute infections), common treatable conditions, and rare disorders (e.g., burning mouth syndrome). |
| E-02 | Effectiveness | Accuracy of Caries Grading and Diagnosis | 2.5/1.75 | 2.5/1 | - | 1 | Graded caries diagnosis is central to operative and endodontic decision-making and prognostication. Diagnostic accuracy is influenced by examination methods, lesion characteristics, operator experience, and adjunctive technologies. Determinants include lesion location and the trajectory of caries progression. |
| E-03 | Effectiveness | Diagnosis of Pulpal Diseases | 3.5/2.5 | 4/0.75 | - | 4 | Diagnosis of pulpal disease is foundational to endodontic care and requires correlating symptom analysis, objective testing, and imaging to determine pulpal status (reversible/irreversible pulpitis, pulpal necrosis, or calcific changes) and guide therapy. |
| E-04 | Effectiveness | Assessment of Periodontal Disease Severity | 3.5/1 | - | - | 3 | Severity assessment in periodontal disease is pivotal to care planning and requires integration of clinical parameters (probing depth, clinical attachment loss, bleeding on probing, tooth mobility), imaging, and systemic modifiers to establish staging and grading and to formulate individualized treatment strategies. |
| E-05 | Effectiveness | Differential Diagnosis of Oral Mucosal Diseases | 3/0.75 | - | - | 2 | Differential diagnosis of oral mucosal diseases requires integration of history, clinical features, laboratory testing, and pathology to distinguish infectious, immune-mediated, neoplastic, and systemic disease manifestations in the oral cavity. |
| E-06 | Effectiveness | Early Clinical Indicators of Oral Neoplasms | 4/0.75 | - | - | 4 | Early clinical indicators of oral malignancy include non-healing ulcers, leukoplakia, erythroplakia, unexplained masses or induration, numbness or paresthesia, tooth mobility or migration, and cervical lymphadenopathy. Site predilection carries distinct oncologic risks, necessitating early recognition and intervention. |
| E-07 | Effectiveness | Diagnosis of Cystic Lesions in the Jaws | 3/0.75 | - | - | 2 | Cystic lesions of the jaws are common in oral and maxillofacial surgery and encompass developmental, inflammatory, and neoplastic entities. Diagnosis requires correlation of clinical presentation, radiographic features, and histopathology to achieve precise classification and individualized treatment. |
| E-08 | Effectiveness | Radiographic Assessment for Impacted Tooth Extraction | 4.5/1.75 | 4.5/1 | - | 5 | Imaging assessment for impacted tooth extraction is a core element of surgical planning, using multimodal imaging to determine three-dimensional tooth position, relationships to adjacent anatomical structures, and potential risks, thereby informing a safe and efficient surgical approach. |
| E-09 | Effectiveness | Risk Assessment of Systemic and Oral Disease Interactions | 2/1 | - | - | 1 | The bidirectional relationship between systemic diseases and oral conditions is central to interdisciplinary care. Dual-risk assessment should elucidate dynamic interactions to mitigate treatment risk and optimize outcomes, with typical examples including diabetes, cardiovascular disease, immunosuppression, hematologic disorders, and chronic kidney disease. |
| E-10 | Effectiveness | Selection of Endodontic Treatment Modalities | 4/1.5 | 4/0 | - | 4 | Endodontic treatment planning should be based on pulpal vitality, extent of disease, systemic health, and functional prognosis of the tooth. |
| E-11 | Effectiveness | Decision for Tooth Preservation versus Extraction | 4/1.5 | 4.5/1 | - | 5 | Decisions between tooth preservation and extraction with subsequent replacement in dental care are complex and consequential, requiring multidimensional evaluation to balance biologic considerations, functional demands, and patient preferences. |
| E-12 | Effectiveness | Selection for Prosthodontic Treatment Plan | 3.5/1 | - | - | 3 | Selection of prosthodontic treatment should be informed by biologic principles, functional requirements, and individualized patient factors, using a multidimensional assessment to craft a precise care plan. |
| E-13 | Effectiveness | Orthodontic Treatment Planning | 3.5/1 | - | - | 3 | Orthodontic treatment planning is the core step in orthodontic care, requiring multidimensional evaluation and systematic analysis to balance functional rehabilitation, esthetic enhancement, and long-term stability. |
| E-14 | Effectiveness | Preventive Care Recommendations in Dentistry | 3.5/1.75 | 3/0 | - | 2 | Prevention of oral diseases requires a systematic strategy that tailors tiered interventions to individual risk profiles, disease types, and population characteristics, with high-risk groups including children, pregnant patients, individuals with diabetes, and patients with a history of head-and-neck radiotherapy. |
| E-15 | Effectiveness | Bone Volume Assessment for Dental Implants | 5/0.75 | - | - | 5 | Alveolar bone volume assessment is central to preoperative implant planning, integrating imaging and clinical examination to evaluate three-dimensional bone quantity (height, width, density) and anatomic constraints (e.g., maxillary sinus, mandibular canal) to guide implant selection and placement strategy. |
| E-16 | Effectiveness | Decision-Making for Timing of Restorative Treatment | 3.5/1 | - | - | 3 | Timing decisions for prosthodontic rehabilitation are a core challenge requiring comprehensive evaluation of systemic status, local tissue stability, and functional needs to balance the risks of "too early" (higher failure risk) versus "too late" (reduced quality of life), achieving durable biologic, mechanical, and esthetic outcomes. |
| E-17 | Effectiveness | Oral Hygiene Instruction and Follow-up Management | 3/1.5 | 3/0 | - | 2 | Oral hygiene instruction and follow-up management are central to prevention and therapy, integrating individualized risk assessment, behavioral science, and digital tools to establish a closed-loop "education–implementation–monitoring" system for sustained oral health. |
| E-18 | Effectiveness | Effective Patient Communication and Informed Consent Essentials | 4/1.5 | 4/0.75 | - | 4 | Effective patient communication and informed consent are core to dental care, directly impacting outcomes, satisfaction, and medicolegal risk. |
| E-19 | Effectiveness | Multidisciplinary Collaboration and Referral Timing in Dental Care | 3/2.25 | 3/0 | - | 2 | Judicious use of multidisciplinary collaboration and timely referral is a key strategy for complex dental cases, integrating specialty expertise to avoid blind spots and provide systemic oral–systemic care—for example, complex implant rehabilitation, combined orthodontic–orthognathic therapy, systemic disease-related oral management, and comprehensive oral oncology care. |
| E-20 | Effectiveness | Behavioral Management in Special Populations | 3/1 | - | - | 2 | Behavioral management for special populations (children, older adults, persons with disabilities, and anxious patients) should integrate physiologic, psychological, and social factors to ensure procedural safety and patient cooperation. |
| E-21 | Effectiveness | Guidance for Systemic Disease Management in Dental Care | 3/2 | 3/1.5 | 3/0 | 2 | Dental care for patients with systemic disease should be guided by dynamic assessment and precise control of overall health to balance procedural safety and long-term outcomes, including patients with diabetes, cardiovascular disease, anticoagulation, chronic kidney disease, and immunosuppression. |

- The scale delineates the consequence of agent error: scores in [0, 3) were mapped to Weight 1 (Context-dependent/limited risk); [3, 3.5)

to Weight 2 (Foundational/low-to-moderate risk); [3.5, 4) to Weight 3 (Moderate risk with meaningful impact); [4, 4.5) to Weight 4 (High-risk and time-sensitive); and [4.5, 5] to Weight 5 (Life-threatening or catastrophic failure).

- S-xx and E-xx denote metrics in the Safety Gate and Effectiveness Gate, respectively.

eTable 3. Test set with structured evaluation suite

| Test set | Bilingual repository source | Checkpoint instances | Rubric points |
|---|---|---|---|
| MCQs | 2,192 English, 2,066 Chinese | - | - |
| Open-end QA (Guideline based) | 19 Chinese, 41 English | 443 | 1,225 |
| Case vignette | 68 Chinese, 50 English | 498 | 1,318 |

Supplementary Note 1. Prompts for Multi-turn Dialogue Simulation

Simulated Patient (SP) prompt:

> You are a patient.
>
> You do not have any medical knowledge.
>
> You have to describe your symptoms from the given case vignette based on the questions asked.
>
> Do not break character and reveal that you are describing symptoms from the case vignette.
>
> Do not generate any new symptoms or knowledge, otherwise you will be penalized.
>
> Do not reveal more information than what the question asks.
>
> Keep your answer short, to only 1 sentence.
>
> Simplify terminology used in the given paragraph to layman language.
>
> **Case Vignette**: {{ case }}

Target LLM prompt:

> You are an AI doctor specializing in {{ specialty }}.
>
> Arrive at a diagnosis of a patient's medical condition.
>
> Ask only one question at a time, and it should not be more than 1 line.
>
> Continue asking questions until you're 100% confident of the diagnosis.
>
> Do not ask the same question multiple times. Ask different questions to cover more information.
>
> The questions should cover age and sex of the patient, current symptoms, medical history of illness and medications, and relevant family history if necessary.
>
> Keep your questions short and brief to not confuse the patient.
>
> After completing all questions, provide a comprehensive plan including assessment, treatment decisions, and patient management.
>
> Your response must begin with "**Final Treatment Plan:**".

eFigure 1. Assessment of Rationale Generation Quality via ROUGE-L Metric.

This figure illustrates the ROUGE-L scores for the rationales generated by each model during the Standardized Exam task. The metric evaluates the linguistic overlap and coherence against reference explanations.

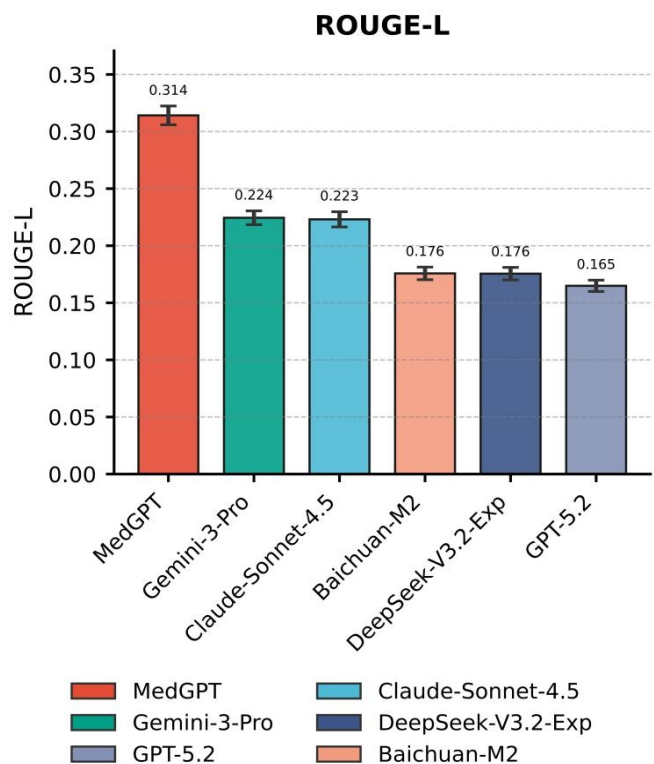

eFigure 2. Worst@k reliability profiling across tasks (MCQs, Open QA, and Multi-turn Dialogue).

This figure quantifies model stability across distinct clinical tasks:

(Left) MCQs - Accuracy: Models exhibit relative stability in static multiple-choice questions, with a gentle decline in the performance floor as k increases.

(Middle) Open QA (Guideline): Stochasticity becomes apparent in guideline-based Q&A, evidenced by a distinct drop in scores.

(Right) Multi-turn Dialogue (Clinical Case): The curves display the steepest deterioration trend in dynamic multi-turn dialogues, indicating significant instability in long-horizon state tracking and information gathering.

Overall, MedGPT (red line) and Gemini-3-Pro (green line) maintain a higher performance floor across tasks, whereas other models exhibit a more rapid performance collapse.

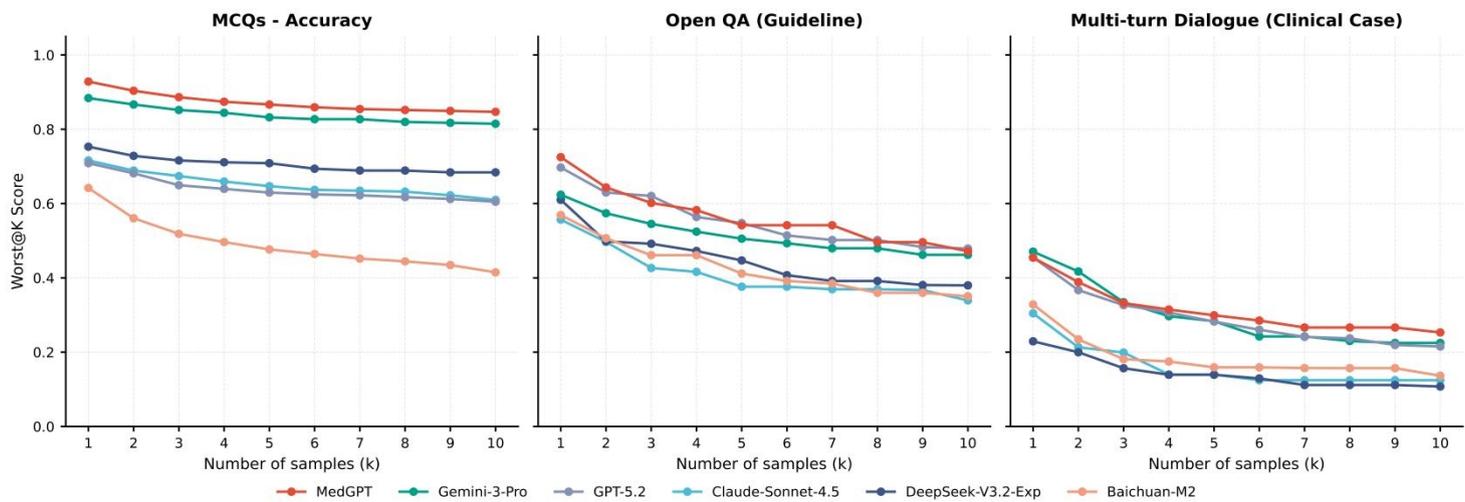

eFigure 3. Assessment of Cross-Lingual Robustness on Natively Sourced Datasets.

Performance comparison stratified by language (English vs. Chinese) across three task types. The bars represent mean scores with 95% confidence intervals. The variation in performance highlights the models' differing capabilities in adapting to distinct linguistic-cultural medical environments.

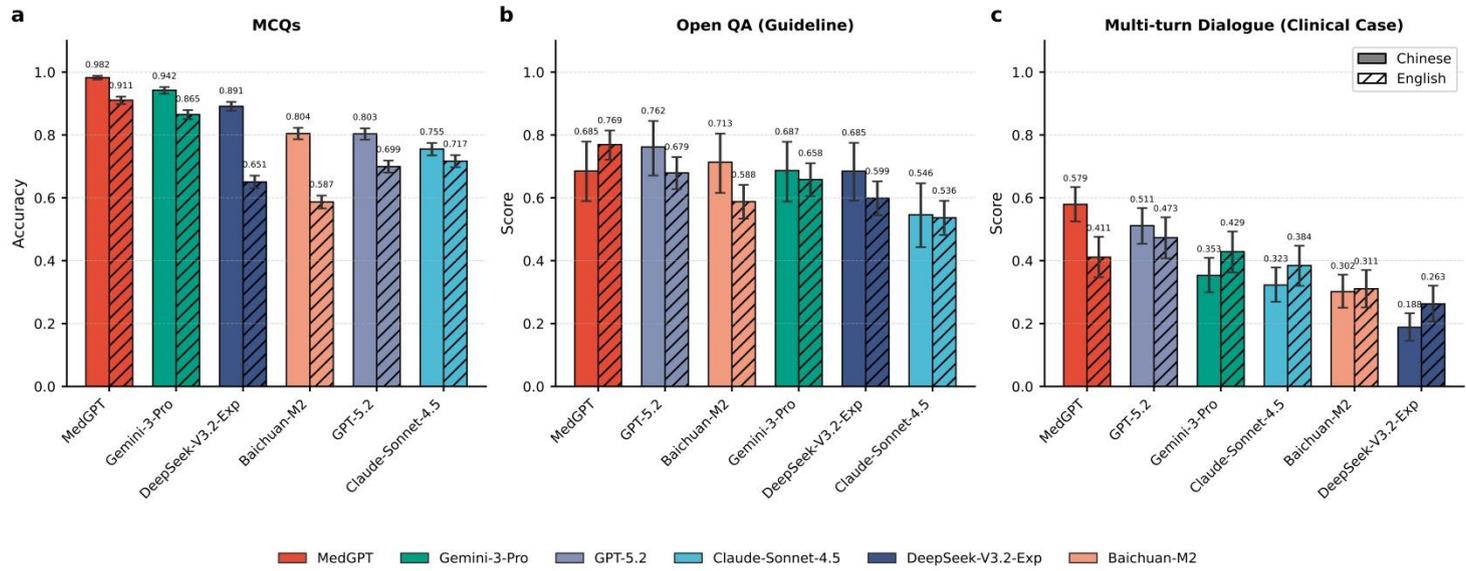